\newenvironment{IEEEkeywords}
{\par\noindent\textbf{Keywords—}\ }
{\par}

\documentclass[letterpaper, 10 pt, conference]{ieeeconf}  % Comment this line out if you need a4paper

\IEEEoverridecommandlockouts                              % This command is only needed if 
\usepackage{graphicx}
\usepackage{booktabs}
\usepackage{amssymb}
\usepackage{hyperref}
\usepackage{graphics} % for pdf, bitmapped graphics files
\usepackage{epsfig} % for postscript graphics files
\usepackage{mathptmx} % assumes new font selection scheme installed
\usepackage{times} % assumes new font selection scheme installed
\usepackage{amsmath} % assumes amsmath package installed

\title{\LARGE \bf
YOLOBirDrone: Dataset for Bird vs Drone Detection and Classification and a YOLO based enhanced learning architecture 
}

\author{Dapinder Kaur$^{1}$ IEEE Member, Neeraj Battish$^{2}$, Arnav Bhavsar$^{4}$, Shashi Poddar$^{3}$ % <-this % stops a space
\thanks{*This work was supported by CSIR-CSIO}% <-this % stops a space
\thanks{$^{1}$Dapinder Kaur is a PhD Scholar with AcSIR, at CSIR CSIO, Chandigarh, 160030, India
        {\tt\small dapinderk.2408@gmail.com}}%
\thanks{$^{2}$Neeraj Battish is a Senior Project Associate with the Department of Intelligent Machine and Computing System, CSIR CSIO, Chandigarh, 160030, India 
        {\tt\small neerajbattish@yahoo.com}}%
\thanks{$^{3}$Shashi Poddar is Senior Principal Principal Scientist with the Department of Intelligent Machine and Computing System, CSIR CSIO, Chandigarh, 160030, India 
        {\tt\small shashipoddar.csio@csir.res.in}}%
\thanks{$^{4}$Arnav Bhavsar is a Associate Professor with School of Computing and Electrical Engineering, IIT Mandi, Himachal Pradesh, 175005, India 
        {\tt\small arnav@iitmandi.ac.in}}%
\thanks{Corresponding Author: Shashi Poddar}
}

\begin{document}

\maketitle
\thispagestyle{empty}
\pagestyle{empty}

%%%%%%%%%%%%%%%%%%%%%%%%%%%%%%%%%%%%%%%%%%%%%%%%%%%%%%%%%%%%%%%%%%%%%%%%%%%%%%%%
\begin{abstract}

The use of aerial drones for commercial and defense applications has benefited in many ways and is therefore utilized in several different application domains. However, they are also increasingly used for targeted attacks, posing a significant safety challenge and necessitating the development of drone detection systems. Vision-based drone detection systems currently have an accuracy limitation and struggle to distinguish between drones and birds, particularly when the birds are small in size. This research work proposes a novel YOLOBirDrone architecture that improves the detection and classification accuracy of birds and drones. YOLOBirDrone has different components, including an adaptive and extended layer aggregation (AELAN), a multi-scale progressive dual attention module (MPDA), and a reverse MPDA (RMPDA) to preserve shape information and enrich features with local and global spatial and channel information. A large-scale dataset, BirDrone, is also introduced in this article, which includes small and challenging objects for robust aerial object identification. Experimental results demonstrate an improvement in performance metrics through the proposed YOLOBirDrone architecture compared to other state-of-the-art algorithms, with detection accuracy reaching approximately 85$\%$ across various scenarios.

\end{abstract}

\begin{IEEEkeywords}
Drone vs. Bird Classification, YOLO, Deep learning, Small object detection, Computer vision
\end{IEEEkeywords}

%%%%%%%%%%%%%%%%%%%%%%%%%%%%%%%%%%%%%%%%%%%%%%%%%%%%%%%%%%%%%%%%%%%%%%%%%%%%%%%%
\section{INTRODUCTION}

Unmanned aerial vehicles (UAVs), also known as drones, have gained immense popularity in various application areas, including surveillance, defense, and environmental protection \cite{Mohsan2022}. Although drones offer several benefits, some unauthorized activities, such as drones carrying explosive payloads or capturing audiovisual data from restricted areas, have also been reported. These safety and privacy issues can cause potential threats that need active monitoring for the detection and identification of unauthorized drones. The rise of drones also raises some concerns about air traffic safety, including the potential for collisions between drones and other aircraft or birds. As a result, various drone detection systems have been developed around the world \cite{Park2021}. However, the detection and tracking of small-sized drones from distant ranges and classifying them from birds in real time using images captured by the visual camera is still an open research problem.

Different deep learning-based approaches have contributed to object detection systems, which are broadly classified into two-stage and one-stage detectors. Two-stage detectors, such as R-CNN, Fast R-CNN, and Faster R-CNN, perform detection in two stages: region proposals and object classification. The first stage extracts the potential object regions, whereas the second stage classifies them and refines the bounding boxes. These detectors are more robust to noise, have higher accuracy, and are better localized. However, their slow convergence and complex structure make them less efficient in real-time detection \cite{Sultana2020}.
On the other hand, one-stage detectors, such as SSD and YOLO architectures, perform detection in a single pass and directly predict bounding boxes and class probabilities, which makes them faster and more straightforward than two-stage detectors. The robustness of the one-stage detectors to scale changes makes them suitable for applications where object sizes vary across different scales. However, the accuracy of one-stage detectors is comparatively lower than that of two-stage detectors. To balance speed and accuracy, different hybrid architectures, attention mechanisms, and feature fusion strategies \cite{Zou2023} have been introduced for efficient detection in real-time system applications.

The task of bird versus drone detection is the most challenging scenario for a detection system due to their varying shapes, sizes, distances, and motions, among other factors. Recently, YOLOv9 was improved by incorporating a modified attention and feature fusion module to classify bird and drone targets using visible and infrared imagery \cite{suo2025yolov9}. In the Drone vs Bird Challenge \cite{Coluccia2024}, a shallow feature pyramid network with attention (SFA) model, used in a single-shot detector (SSD), adaptively adjusts the layers and fuses features to extract deep semantic information. This approach detects drones in 11 out of 30 videos, achieving an overall score of 0.12 \cite{Dong2023}. On the other hand, in the same challenge, YOLOv7 with channel and spatial reliability tracking utilizes bounding boxes with high confidence values to enhance detection and reduce false detections \cite{Mistry2023}. Another drone vs. bird challenge \cite{Coluccia2021} presented different deep learning models, and YOLOv5, combined with synthetic data generation and a tracking-based method, ranked first with an overall precision of 0.677. These challenges focus on detecting and classifying drones from birds. However, the detection of birds was not included in this challenge. Inception-v3-CNN \cite{Torky2024}, EfficientNetB6 \cite{Ghazlane2024}, YOLOv4, YOLOv5 \cite{SethuSelvi2022}, YOLOv5m6 \cite{Coluccia2024} etc. are the recent developments in deep learning architecture which were designed and tested on a benchmarked dataset of drones and birds, and perform better in compare to some standard architectures such as VGG, ResNet, Inception, DenseNet and Xception architectures. However, high computational complexity and limited datasets can challenge the robustness of these architectures. It is therefore proposed here to study and propose a scheme that can simultaneously detect and classify drones and birds, thereby improving overall precision and accuracy. This will help bridge the gap between the requirements for intricate solutions in safety applications and create a roadmap toward addressing current difficulties.

The main contribution of this paper is a novel YOLOBirDrone architecture for the accurate detection and classification of drones and birds. The novelty of this proposed architecture includes:
\begin{itemize}
\item An adaptive and extended layer aggregation network (AELAN) based architecture that can adapt to the object shape and improve feature localization for bird vs. drone detection and classification.
\item Multi-scale progressive dual attention module (MPDA) and reverse MPDA (RMPDA) to enhance the spatial and channel information at multiple stages and extract multi-scale features for small object detection.
\item Model training using a large-scale self-generated dataset (BirDrone) and self-annotated video frames from the drone vs bird detection benchmarked dataset for detection as well as classification tasks.
\end{itemize}
The next section of the paper, Section 2, discusses the literature, including various architectures and approaches that inform the proposed methodology of this work, presented in Section 3. The experimentation and evaluation of the results are then described both quantitatively and qualitatively, along with the comparative analysis in Section 4. Finally, the conclusion and future scope are presented in Section 5.

\section{RELATED WORK}
The research work focuses on one-stage detectors for real-time detection and classification systems, proposing a novel YOLO architecture. In this section, an overview of different YOLO architectures is presented first. Secondly, the attention and their different approaches are delineated. Overall, this related work provides background information that is used to understand the requirements for designing a novel architecture for the proposed research work.

\subsection{YOLO-based architectures}
You Only Look Once (YOLO) is a one-stage detector that performs detection and classification in a single pass by utilizing a grid-based method, thereby improving speed for real-time detections. This architecture mainly consists of three modules: (i) Backbone- to extract features at different pyramid levels, (ii) Neck- to fuse the features at different pyramid levels, and (iii) Head- to detect and classify the object(s). Several versions of YOLO have been developed so far, with the focus on improving the models' capabilities to meet specific requirements. The first version of YOLO, YOLOv1, can predict only two boxes of the same class and faces several difficulties during object detection. YOLOv2 enhances feature extraction to improve detection. Still, it fails to detect small objects, has lower localization accuracy, and suffers from class imbalances. By adding feature pyramid networks, YOLOv3 improves the low-level features. It adds multi-scale capabilities but still struggles with tiny objects, is inconsistent, and is computationally expensive. YOLOv4 solves the data imbalance issue and improves both the speed and accuracy of the model. High computational complexity, suboptimal detection, dependency on anchor boxes, and increased model size are some of its drawbacks \cite{Diwan2023}. Different modifications make the YOLOv5 architecture a lighter, faster, and simpler version of YOLO, which improves small object detection, accuracy, and speed, among other aspects.
Like YOLOv4, YOLOv5 also relies on anchor boxes, which can sometimes negatively impact performance and offer limited support for complex tasks. Unlike other models, YOLOv6 does not have a DarkNet-based backbone. It utilizes reparameterized VGG for feature extraction, which enhances its speed. However, it is still not capable of dealing with small object detection problems and has a complex architectural structure \cite{Terven2023}. YOLOv7, YOLOv8, and YOLOv9 have modified the architectural structures of YOLOv5 to enhance the model's capabilities. YOLOv10, YOLOv11, and YOLOv12 further advance the YOLO series by introducing optimized backbone and neck architectures, enhanced feature fusion strategies, and improved training techniques, resulting in higher detection accuracy, better small-object localization, and more efficient inference for real-time applications. All these models have their own strengths but still lack when used for small object detection tasks in complex environments \cite{Alif2024}. Several other advancements have been made in these architectures, including approaches such as scaling, feature fusion, context modeling, feature imitation, and attention methods \cite{Cheng2023}. However, there is scope for improvement, specifically in small object detection and classification tasks, such as bird versus drone detection and classification. This research work primarily focuses on an attention-based approach; therefore, some of the existing attention mechanisms are discussed in the subsequent section. 

\subsection{Attention mechanisms}

Attention mechanisms are a popular approach that was initially introduced in natural language processing (NLP), where they utilize relevant information to retain longer sequences and enhance the performance of various NLP tasks \cite{Soydaner2022}. These mechanisms have now become the most important concept in deep learning. They are also applied in various computer vision tasks, such as object detection, image captioning, and segmentation, among others \cite{Brauwers2023}. In computer vision tasks, attention is primarily categorized into spatial and channel attention, where spatial attention focuses on the location of relevant information, and channel attention emphasizes the type of relevant information present. Several different attention approaches have been designed to date, utilizing various operations and methods, including pooling, fully connected layers, concatenation, and convolution. In this section, some of the recently developed attention mechanisms for enhancing YOLO architectures are discussed. Squeeze-and-Excitation (SE) attention, which primarily focuses on spatial information, was utilized to enhance the YOLOv5 architecture \cite{Lv2024}. However, it sometimes ignores the importance of spatial information, which can lead to false detections \cite{Zhang2023}. The channel and multi-scale attention were also utilized to enhance the YOLOv5 architecture. Although these attention mechanisms efficiently extract channel information and handle multi-scale objects, their complexity and limited flexibility are drawbacks, making them less effective in complex scenarios \cite{Qing2024}. A novel concept of detach and merge attention was designed, which implements both spatial and channel attention to enhance the model’s capabilities but is computationally expensive in nature \cite{Li2024}. Nowadays, transformer-attention approaches are incorporated as self-attention modules along with spatial and channel attention, such as convolutional block attention modules (CBAM). Still, the risk of overfitting, computational cost, and other limitations are associated with such methods \cite{Zedda2024}. Based on this and several other attention-based architectures, not limited to \cite{Kang2024} and \cite{Sekharamantry2024}, it has been identified that there is a need for an attention approach that can capture relevant details while also requiring computational efficiency to prevent increasing model complexity.\\
Therefore, this research focuses on the recently developed YOLOv9 architecture and aims to propose different novel concepts to enhance its performance for the task of drone vs bird detection and classification. 

\section{PROPOSED METHODOLOGY}
This work proposes an improved YOLO architecture, named YOLOBirDrone, which utilizes YOLOv9 as its base architecture and enhances its performance for the detection and classification of Drones and Birds. This proposed YOLOBirDrone consists of different modules, including (a) AELAN, (b) MPDA, and (c) RMPDA attention approaches embedded in the backbone and neck of the model to improve its overall efficiency. The baseline and proposed architectures are discussed in the subsequent sub-sections.
\subsection{YOLOv9}
YOLOv9 is a recent development of the YOLO architecture, which ensures that no amount of information is ignored in establishing correct associations between inputs and targets to enhance predictions. In this architecture, a generalized efficient layer aggregation network (GELAN) and the concept of programmable gradient information (PGI) were introduced to obtain complete information and achieve better results. In this, GELAN combines two different architectures: cross-stage partial networks (CSPNet), which were introduced in earlier YOLO versions, and an efficient layer aggregation network (ELAN) from YOLOv7. In YOLOv9, instead of stacking convolution layers like in YOLOv7, any computational blocks can be incorporated. By experimentation with convolution, ResNet, DarkNet, and CSPNet blocks, it has been found that the performance of CSPNet is better with ELAN and CSP depth of \{2,3\}.\\
On the other hand, PGI has three main components: the main branch, the auxiliary reversible branch, and the multi-level auxiliary information. The main branch is a simple neck-to-head connection that fuses low and high-level features to form accurate predictions. The auxiliary reversible branch adds additional connections from deep to shallow layers to reduce information loss, and multi-level auxiliary information aggregates gradient information at deeper levels. However, it can increase the computational cost; for this reason, these branches are included during model training, not during inference. Overall, this recent version of YOLO features a strong and stable architecture, resulting in a significant improvement in the model's accuracy for the object detection task \cite{Wang2024}.
\subsection{YOLOBirDrone}
\begin{figure}[thpb]
      \centering
     \includegraphics[scale=0.25]{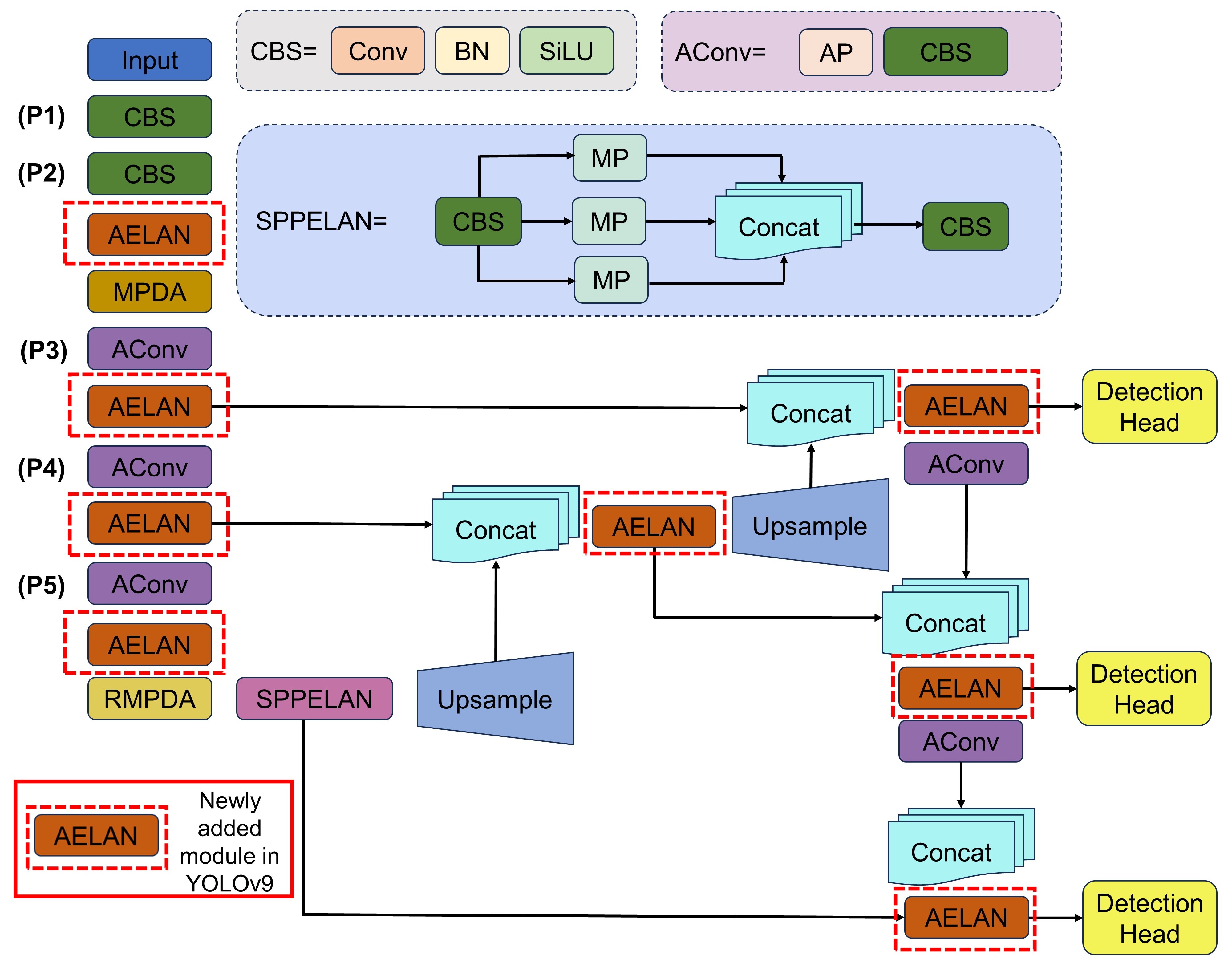}  
      \caption{Proposed YOLOBirDrone Architecture}
      \label{figurelabel:figure1}
   \end{figure}

This research aims to develop a novel YOLO architecture with enhanced capabilities for detecting and classifying drones and birds. The task of bird vs. drone detection and classification is more challenging than other object detection problems due to the objects' similar properties, such as size, shape, and motion. It is even more complex when both the drones and birds are far from the camera. Considering these complexities, a YOLOBirDrone architecture is proposed, which is an improved version of YOLOv9 using different modules: AELAN, MPDA, and RMPDA. Using these modules, the backbone and neck of the YOLOv9 architecture enrich the feature information for aggregation and predictions. The overall architecture of YOLOBirDrone is presented in Figure~\ref{figurelabel:figure1}, and its details are discussed in the following subsections.
\subsubsection{AELAN}
\begin{figure}[thpb]
      \centering
     \includegraphics[scale=0.25]{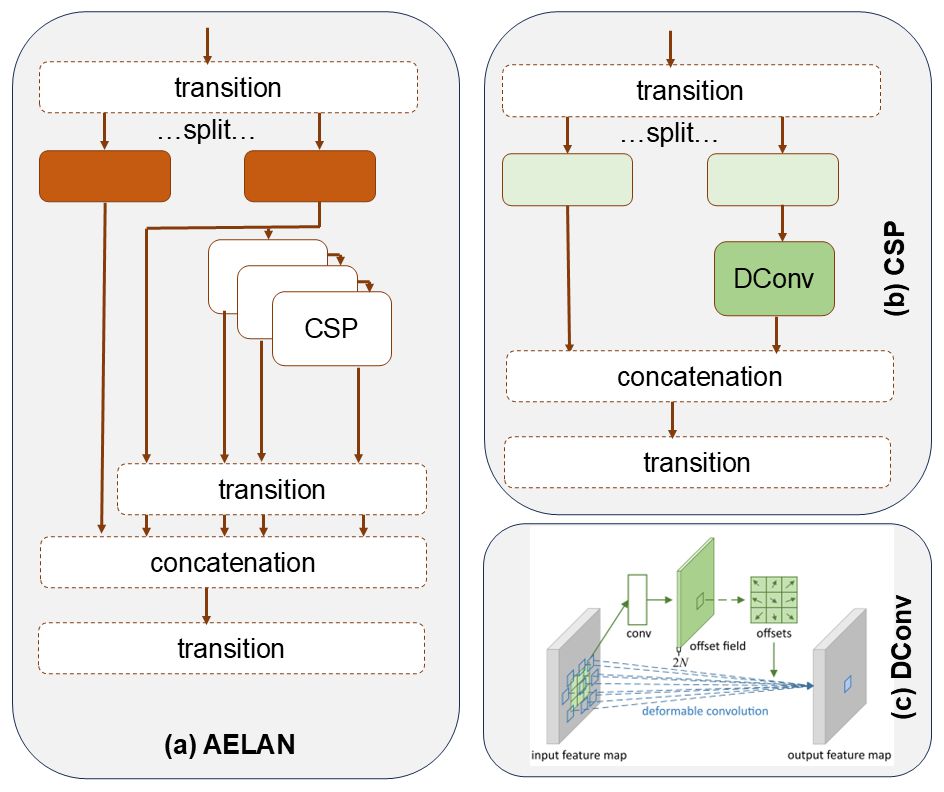}
      
      \caption{Proposed AELAN sub-architecture}
      \label{figurelabel:figure2}
   \end{figure}

The proposed AELAN is an extension of GELAN, as shown in Figure~\ref{figurelabel:figure2}, which has been improved using the concept of deformable convolutions (DConv) \cite{Dai2017}. GELAN is a combination of ELAN and CSPNet, and instead of using standard convolution layers in CSPNet, AELAN introduced DConv layers. DConv is an advanced convolution layer that allows the sampling grid of the convolution to shift based on the input, enabling the kernel to adapt to the receptive field. The output at position x is
\begin{equation}
y(x) = \sum_{k=1}^{K} w_k \cdot x(x + p_k + \Delta p_k)
\end{equation}
Where \( \Delta p_k \) is the learnable offsets. These learnable offsets for each position shift sampling points to more relevant locations within an image and are learned through separate convolutional layers. The offsets can be obtained using
\begin{equation}
\Delta p = f_{\theta}(x)
\end{equation}
Here, \( f_{\theta} \) is the offset learning module with parameter \( \theta \). Therefore, it enables the kernel to adapt to the shape and size of the object’s spatial distribution, provides a flexible receptive field, and also facilitates the extraction of feature information for small and irregular objects. This is the main reason for using DConv layers instead of standard convolutions. It also offers better localization and handles scale variations by shifting the receptive field of the kernel. Overall, its greater flexibility, improved feature extraction, and adaptability to the shape of the object benefit the model architecture in extracting relevant information about the object and better classifying drones and birds.

\subsubsection{MPDA and RMPDA}
\begin{figure*}[t]
    \centering
    \includegraphics[width=0.7\textwidth]{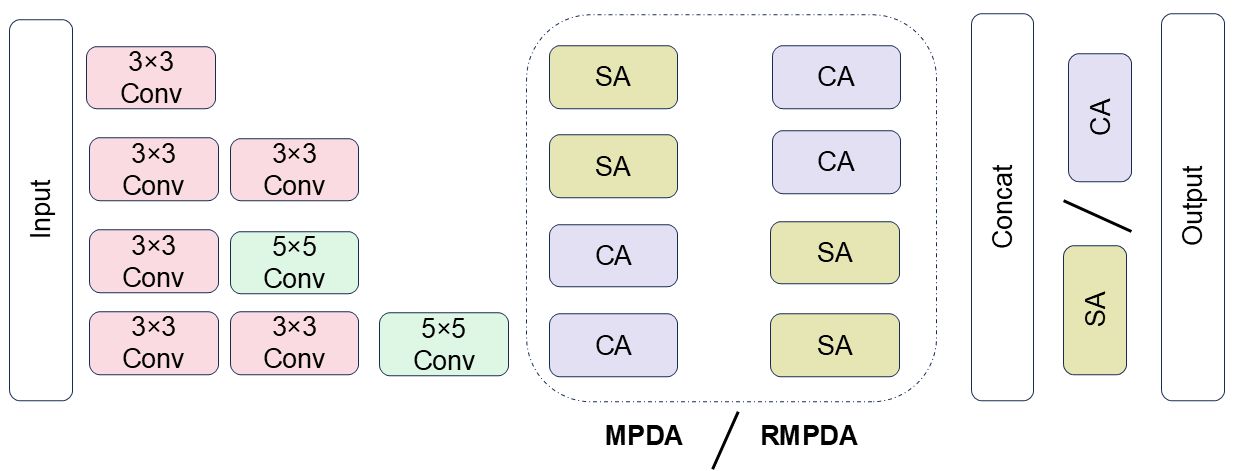}
    \caption{MPDA and RMPDA Modules}
    \label{figurelabel3}
\end{figure*}

The idea of MPDA originates from three different concepts$:$ multi-scaling \cite{Battish2024}, Progressive approach \cite{Wang2024a}, and Dual attention \cite{Fu2019}. As the research task involves variable-sized objects, the multiscale capabilities can accommodate the scale variations. The progressive approach, also known as a multi-staged approach, refines the feature or attention maps across several layers, thereby improving the model's ability to focus on fine-grained details. On the other hand, dual attention here implies the aggregation of spatial and channel attention from different receptive fields. Figure ~\ref{figurelabel3} presents the MPDA and RMPDA approach, where features of four different scales are captured using different combinations of 3×3 and 5×5 kernel and results in the receptive field equivalent to 3×3(\( X_1 \)), 5×5\( (X_2) \), 7×7\( (X_3) \), and 9×9\((X_4) \). The combination of small-sized kernels, specifically 3×3 and 5×5 kernels, reduces the number of trainable parameters while maintaining a higher receptive field, which yields better results without overloading the model. Assume the input feature map is \( X \in \mathbb{R}^{H \times W \times C} \), where H, W, and C are the height, width, and channels. The multiscale features are obtained using different convolutions, with each scale corresponding to a distinct receptive field. The multiscale features for the given input are:
\begin{equation}
X_1 = f_{3 \times 3}(X), \, X \in \mathbb{R}^{H \times W \times C_1}
\end{equation}
\begin{equation}
X_2 = f_{3 \times 3}(f_{3 \times 3}(X)), \, X \in \mathbb{R}^{H \times W \times C_2}
\end{equation}
\begin{equation}
X_3 = f_{5 \times 5}(f_{3 \times 3}(X)), \, X \in \mathbb{R}^{H \times W \times C_3}
\end{equation}
\begin{equation}
X_4 = f_{5 \times 5}(f_{3 \times 3}(f_{3 \times 3}(X))), \, X \in \mathbb{R}^{H \times W \times C_4} 
\end{equation}
The attention is then applied to each \( X_i \) scale, generating different feature maps. If \( A(\cdot) \) represent the attention, then the enhanced feature Maps \( X_i' \) computed as:
\begin{equation}
    X_i' = A(X_i) \odot X_i
\end{equation}
After this step, these feature maps are concatenated, and attention is applied to the concatenated output, as given in equation (8).
\begin{equation}
F_{\text{final}} = A(\text{Concat}(X_1, X_2, X_3, X_4)) \odot (\text{Concat}(X_1, X_2, X_3, X_4))
\end{equation}
Dual attention at different scales leverages both local and global context information, thereby improving the model's effectiveness. The difference between MPDA and RMPDA is the spatial and channel attention computation on different scales, as well as after aggregation. As shown in Figure~\ref{figurelabel:figure1}, MPDA is embedded at initial pyramid levels with high resolution and fewer channels, whereas RMPDA is applied at deeper levels with lower resolution and more channels. As initial pyramid levels contain more spatial information due to high resolution, when SA is applied on \( (X_1) \) and \( (X_2) \) scales, it can focus more on fine-grained details as well as local features and enhance the model capabilities to detect small objects. On the other hand, when CA is applied at this level on \( (X_3) \) and \( (X_4) \), it refines the important patterns and prioritizes the global information to understand the complex scenes. This case is reversed at more profound levels because of the higher number of channels. In this work, EPSANet \cite{Kaur2024} is employed for spatial attention, and EECA \cite{Liu2022} is utilized for channel attention.

\section{EXPERIMENTATION AND RESULTS ANALYSIS}
The proposed YOLOBirDrone architecture is implemented using the Python platform, and all the experiments are performed on a Windows-based system with an NVIDIA RTX graphics card. The proposed model is trained with an input size of 640×640, a batch size of 16, 300 epochs, and a learning rate of 0.01. The following sub-sections discuss the dataset, ablation study, and qualitative and quantitative results.

\subsection{Dataset details} 
A large-scale Drone-Bird detection dataset, BirDrone, is introduced in this work to support research on reliable discrimination between drones and birds in various environments. It is designed to address the challenges of visual similarity, scale variation, background clutter, and motion ambiguity that commonly arise in real-world aerial surveillance scenarios. The dataset was collected in the Green Zone area of Chandigarh, India, using consumer-grade acquisition devices, including a webcam and a mobile phone camera, to reflect realistic and low-cost surveillance conditions. The image dataset comprises 8,428 high-resolution images that contain two visually confusable object categories: drones and birds, which appear at different altitudes, orientations, and distances from the camera. A key characteristic of the dataset is the presence of small-sized objects and low-contrast objects, which reflects practical surveillance conditions where targets often occupy a limited number of pixels. The dataset also includes challenging cases such as partial occlusions, motion blur, and complex background elements, which increase the difficulty in accurate classification and localization. Additionally, drones and birds appear at arbitrary spatial locations, including near image boundaries, which increases detection difficulty and encourages spatial robustness in detection models. The diversity in the dataset ensures robustness against overfitting and enhances generalization across different scenarios. 

Each image is manually annotated with precise bounding box labels following standard object detection protocols. The annotations include object class labels, i.e., drones or birds, along with their corresponding spatial coordinates. The labelling process was carefully validated to ensure annotation consistency and accuracy, making it suitable for supervised learning, benchmarking, and comparative evaluation of detection models. The BirDrone dataset is intended to facilitate the development and evaluation of a robust drone-bird classification algorithm for applications such as airspace security, wildlife monitoring, and counter-UAV systems. By providing a large number of challenging and diverse samples, the dataset helps close the gap between controlled benchmark datasets and real-world aerial detection scenarios. 

Additionally, the drone vs. bird detection challenge dataset \cite{Coluccia2024} is also incorporated in this research work. The bounding box coordinate information for birds is not available in this dataset. So only four videos (3067 frames) are manually extracted and annotated for both birds and drones. In this way, the data under different environmental conditions included in this study adds more generalization to the training model. In total, 11,495 image samples of drones and birds are used for training, testing, and validation, with ratios of 70%, 20%, and 10%, respectively. The other details about the dataset are described in Table~\ref {tab:dataset_details}.

\begin{table}[h!]
\centering
\caption{Dataset Details}
\label{tab:dataset_details}
\small % slightly smaller font to fit column
\setlength{\tabcolsep}{6pt} % reduce horizontal padding
\begin{tabular}{lc}
\toprule
\textbf{Parameter} & \textbf{Value} \\
\midrule
Total Images & 11,495 \\
Drone Objects & 13,881 \\
Bird Objects & 15,867 \\
Smallest Drone & 7$\times$5 pixels \\
Smallest Bird & 6$\times$7 pixels \\
Extremely Small Targets ($<$20$\times$20) & 1,129 \\
Small Targets (20$\times$20$<$ and $<$32$\times$32) & 1,576 \\
Medium Targets (32$\times$32$<$ and $<$96$\times$96) & 12,510 \\
Large Targets ($>$96$\times$96) & 14,553 \\
\bottomrule
\end{tabular}
\end{table}

\subsection{Ablation Study}
The YOLOBirDrone architecture incorporates various modules into the baseline YOLOv9 architecture to enhance the model's capabilities, thereby improving drone vs. bird detection and classification performance. These modules are added to the baseline architecture to yield different models from (M1 to M6), indicated in Table~\ref {Table_1}. In this, M1 represents a baseline YOLOv9 architecture, and M2 incorporates AELAN into the YOLOv9 architecture, enabling the architecture to adapt to object size and shape. M3, M4, and M5 represent the addition of MPDA, RMPDA, and both modules to the baseline YOLOv9 architecture, enhancing global and local context information. M6 incorporates all the proposed modules to enhance the accuracy of detecting and classifying objects of different sizes.

\begin{table}[h!]
\centering
\caption{Different Models Devised for Ablation Study}
\label{Table_1}
\small % slightly smaller font to fit column
\setlength{\tabcolsep}{6pt} % reduce horizontal padding
\begin{tabular}{lccc}
\toprule
\textbf{Models} & \textbf{AELAN} & \textbf{MPDA} & \textbf{RMPDA} \\
\midrule
M1 & $\times$ & $\times$ & $\times$ \\
M2 & $\checkmark$ & $\times$ & $\times$ \\
M3 & $\times$ & $\checkmark$ & $\times$ \\
M4 & $\times$ & $\times$ & $\checkmark$ \\
M5 & $\times$ & $\checkmark$ & $\checkmark$ \\
\textbf{M6} & $\checkmark$ & $\checkmark$ & $\checkmark$ \\
\bottomrule
\end{tabular}
\end{table}

\begin{table}[h!]
\centering
\caption{Performance Evaluation of Different Proposed Models}
\label{tab:performance_metrics}
\small % slightly smaller font to fit column
\setlength{\tabcolsep}{6pt} % reduce horizontal padding
\begin{tabular}{lcccc}
\toprule
\textbf{Models} & \textbf{P} & \textbf{R} & \textbf{mAP$^{0.5}$} & \textbf{mAP$^{0.5-0.95}$} \\ 
\midrule
M1 & 0.929 & 0.907 & 0.940 & 0.644 \\
M2 & 0.928 & 0.911 & 0.939 & 0.639 \\
M3 & 0.931 & 0.911 & 0.938 & 0.629 \\
M4 & 0.933 & 0.909 & 0.939 & 0.639 \\
M5 & 0.931 & 0.909 & 0.938 & 0.638 \\
\textbf{M6} & \textbf{0.949} & \textbf{0.917} & \textbf{0.948} & \textbf{0.668} \\
\bottomrule
\end{tabular}
\end{table}

 The performance metrics, including precision (P), recall (R), $ \text{mAP}^{0.5} $, and $ \text{mAP}^{0.5-0.95} $, as presented in Table~\ref {tab:performance_metrics}, describe the impact of adding modules on their performance. Using AELAN in M2, the model's ability to adapt to the shape of the object and complex spatial patterns has increased. This leads to better localization of objects, generates fewer false positives, misses fewer objects, and hence improves its P, R, and other performance parameters, clearly seen in the results. 
The modules M3, M4, and M5 utilize the attention modules MPDA and RMPDA to capture local and global context information in distinct ways. It is seen that, in most cases, these modules performed better than M2 and M1. This is because the model becomes more selective and focuses on the most important regions and features when spatial and channel attention is aggregated from different receptive fields and applied at different stages. Combining all these modules in M6 collectively enhances the scale, shape, and local and global context information, adding significant capabilities to the model and achieving better results in all aspects. These results are obtained at 300 epochs, and it is clear from the results that module M6 (proposed YOLOBirDrone) achieves higher performance than all other models.

\subsection{Detection Results}
\begin{figure}[thpb]
    \centering
    \includegraphics[width=3in]{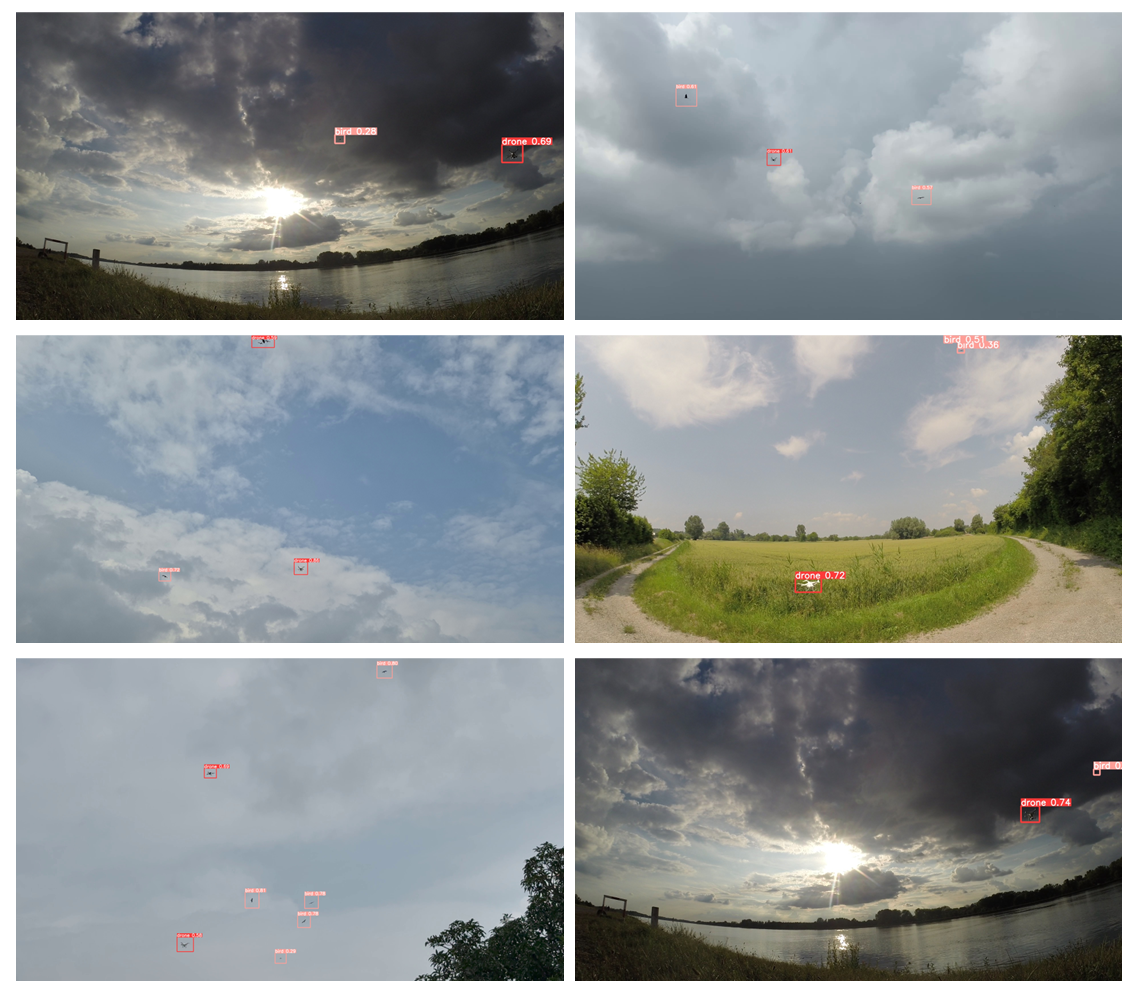} % Change to .png or .jpg
    \caption{Detection results of proposed YOLOBirDrone}
    \label{figurelabel4}
\end{figure}

The visual results of the proposed YOLOBirDrone for the task of birds and drone detection and classification are presented in Figure~\ref {figurelabel4}. These detection results show that the proposed YOLOBirDrone accurately detects and classifies both birds and drones. The detection accuracy of different models, as shown in Table~\ref{tab:detection_accuracy}, also indicates that the proposed model, M6, i.e., YOLOBirDrone, achieves the highest detection accuracy of 84.91\%. Furthermore, it yields a reduced number of false detections, as evidenced by lower false negatives (FN) and false positives (FP) compared to other methods, indicating improved detection reliability.

\begin{table}[h!]
\centering
\caption{Detection Accuracy and False Detections of Different Proposed Models}
\label{tab:detection_accuracy}
\small % slightly smaller font to fit column
\setlength{\tabcolsep}{6pt} % reduce horizontal padding
\begin{tabular}{lccc}
\toprule
\textbf{Models} & \textbf{Detection Accuracy (\%)} & \textbf{FN (\%)} & \textbf{FP (\%)} \\ 
\midrule
M1 & 81.73 & 13.21 & 5.04 \\
M2 & 82.86 & 12.34 & 4.78 \\
M3 & 82.52 & 11.65 & 5.82 \\
M4 & 83.04 & 12.69 & 4.26 \\
M5 & 83.43 & 11.95 & 4.12 \\
\textbf{M6} & \textbf{84.91} & \textbf{11.61} & \textbf{3.73} \\
\bottomrule
\end{tabular}
\end{table}
{\footnotesize*FN denotes instances where an object is present but not detected, and FP denotes instances where a detection is made incorrectly (i.e., no object is present).}

\subsection{Comparison of the proposed YOLOBirDrone with other State-of-the-art (SOTA) methods}

The performance of the proposed YOLOBirDrone architecture in terms of different performance metrics is compared with existing SOTA methods, including different YOLO architectures and transformer-based architecture, RT-DETRv2 \cite{lv2024rt}. According to the literature \cite{Coluccia2021a}, the precision of 0.677 has been achieved with the YOLOv5 and 0.782 mAP with the modified version of YOLOv9 \cite{suo2025yolov9}. This comparison study, however, considers different models and compares the performance of the proposed YOLOBirDrone architecture based on various performance metrics.

\begin{table*}[ht!]
\centering
\caption{Comparison of Proposed YOLOBirDrone with SOTA Methods}
\label{tab:performance_metrics1}
\small % slightly smaller font
\setlength{\tabcolsep}{4pt} % reduce horizontal padding
\begin{tabular}{lcccccc}
\toprule
\textbf{Models} & \textbf{P} & \textbf{R} & \textbf{mAP$^{0.5}$} & \textbf{mAP$^{0.5-0.95}$} & \textbf{Detection Accuracy (\%)} & \textbf{AIT/frame (s)} \\ 
\midrule
YOLOv8 & 0.933 & 0.912 & 0.947 & 0.661 & 81.82 & 0.190 \\
YOLOv9 & 0.929 & 0.907 & 0.940 & 0.644 & 81.73 & 0.211 \\
YOLOv10 & 0.925 & 0.886 & 0.931 & 0.626 & 72.26 & 0.154 \\
YOLOv11 & 0.928 & 0.911 & 0.939 & 0.654 & 81.04 & 0.274 \\
YOLOv12 & 0.939 & 0.888 & 0.931 & 0.636 & 77.39 & 0.274 \\
RT-DETRv2 & 0.929 & 0.876 & 0.938 & 0.633 & 80.24 & 0.275 \\
\textbf{YOLOBirDrone} & \textbf{0.949} & \textbf{0.917} & \textbf{0.948} & \textbf{0.668} & \textbf{84.91} & \textbf{0.149} \\
\bottomrule
\end{tabular}
\end{table*}

Table ~\ref{tab:performance_metrics1} presents that the proposed YOLOBirDrone achieves the highest precision and recall, indicating its ability to avoid more false positives compared to other architectures due to its better object localization. The highest $ \text{mAP}^{0.5} $ and $ \text{mAP}^{0.5-0.95} $ demonstrate that the proposed YOLOBirDrone not only accurately localizes the objects but also maintains consistent performance across different levels of overlap between predicted and ground-truth bounding boxes. In addition, the proposed model attains high detection accuracy and a low average inference time (AIT) per frame, indicating a favourable balance between detection performance and efficiency. In this way, each metric represents the effectiveness of the proposed YOLOBirDrone in the task of detecting and classifying birds and drones.

\section{CONCLUSION}
This research work proposes the YOLOBirDrone architecture for the detection and classification of birds and drones. This proposed architecture incorporated the use of deformable convolution in the backbone and neck of the YOLOv9 architecture. It proposed an AELAN backbone that can adapt to the object shape to improve classification. The modules MDPA and RMPDA progressively refine the attention maps at different stages, focusing on both spatial and channel information to preserve features and enrich the local and global information of objects. The ablation study results present the performance achieved by embedding these modules into the YOLOv9 architecture. It is found that the proposed YOLOBirDrone architecture, when using AELAN with both MPDA and RMPDA, achieves higher performance than others. The dataset used for this evaluation is a combination of a self-generated dataset and samples from an existing benchmarked dataset. The detection accuracy of YOLOBirDrone is also higher than that of the baseline YOLOv9 and other advanced models. The comparison with SOTA methods also states that the proposed architecture has an efficient structure that improves the performance of the bird versus drone detection and classification. 

\section*{Data Availability}
The dataset that supports the findings of this study is preliminarily available at \url{https://github.com/dapinderk-2408/YOLOBirDrone} and the detailed version of the dataset can be requested from the corresponding author.

\addtolength{\textheight}{-12cm}   % This command serves to balance the column lengths

%%%%%%%%%%%%%%%%%%%%%%%%%%%%%%%%%%%%%%%%%%%%%%%%%%%%%%%%%%%%%%%%%%%%%%%%%%%%%%%%

%%%%%%%%%%%%%%%%%%%%%%%%%%%%%%%%%%%%%%%%%%%%%%%%%%%%%%%%%%%%%%%%%%%%%%%%%%%%%%%%

%%%%%%%%%%%%%%%%%%%%%%%%%%%%%%%%%%%%%%%%%%%%%%%%%%%%%%%%%%%%%%%%%%%%%%%%%%%%%%%%

\bibliographystyle{plain}
% Add the bib file
\bibliography{icrarefernce.bib}
\end{document}